\RequirePackage{fix-cm}
\documentclass[twocolumn]{svjour3}          
\smartqed  
\usepackage{amsmath,amssymb} 
\usepackage[dvipdfmx]{graphicx}
\usepackage{subfigure}
\usepackage{wrapfig}
\usepackage{hyperref}
\usepackage{colortbl,array,xcolor}

\makeatletter
\newcommand{\figcaption}[1]{\def\@captype{figure}\caption{#1}}
\newcommand{\tblcaption}[1]{\def\@captype{table}\caption{#1}}
\makeatother

\usepackage{bm}
\usepackage{comment}

\usepackage{color}

\begin{document}

\title{Pre-training without Natural Images
}


\author{Hirokatsu Kataoka \and
        Kazushige Okayasu \and
        Asato Matsumoto \and
        Eisuke Yamagata \and
        Ryosuke Yamada \and
        Nakamasa Inoue \and
        Akio Nakamura \and
        Yutaka Satoh    
}


\institute{H. Kataoka, K. Okayasu, A. Matsumoto, R. Yamada, Y. Satoh \at
              Artificial Intelligence Research Center (AIRC), National Institute of Advanced Industrial Science and Technology (AIST) \\
              \email{hirokatsu.kataoka@aist.go.jp} 
            \and
            E. Yamagata, N. Inoue \at
                Tokyo Institute of Technology
            \and
            A. Nakamura \at
                Tokyo Denki University
}


\maketitle

\begin{abstract}
Is it possible to use convolutional neural networks pre-trained without any natural images to assist natural image understanding? The paper proposes a novel concept, Formula-driven Supervised Learning. We automatically generate image patterns and their category labels by assigning fractals, which are based on a natural law existing in the background knowledge of the real world. Theoretically, the use of automatically generated images instead of natural images in the pre-training phase allows us to generate an infinite scale dataset of labeled images. Although the models pre-trained with the proposed Fractal DataBase (FractalDB), a database without natural images, does not necessarily outperform models pre-trained with human annotated datasets at all settings, we are able to partially surpass the accuracy of ImageNet/Places pre-trained models. The image representation with the proposed FractalDB captures a unique feature in the visualization of convolutional layers and attentions.\footnote{The codes, datasets, and pre-trained models are publicly available: \url{https://github.com/hirokatsukataoka16/FractalDB-Pretrained-ResNet-PyTorch}}

\end{abstract}

\begin{figure*}[t]
  \centering
    \subfigure[The pre-training framework with Fractal geometry for feature representation learning. We can enhance natural image recognition by pre-training without natural images.]{\includegraphics[width=0.54\linewidth]{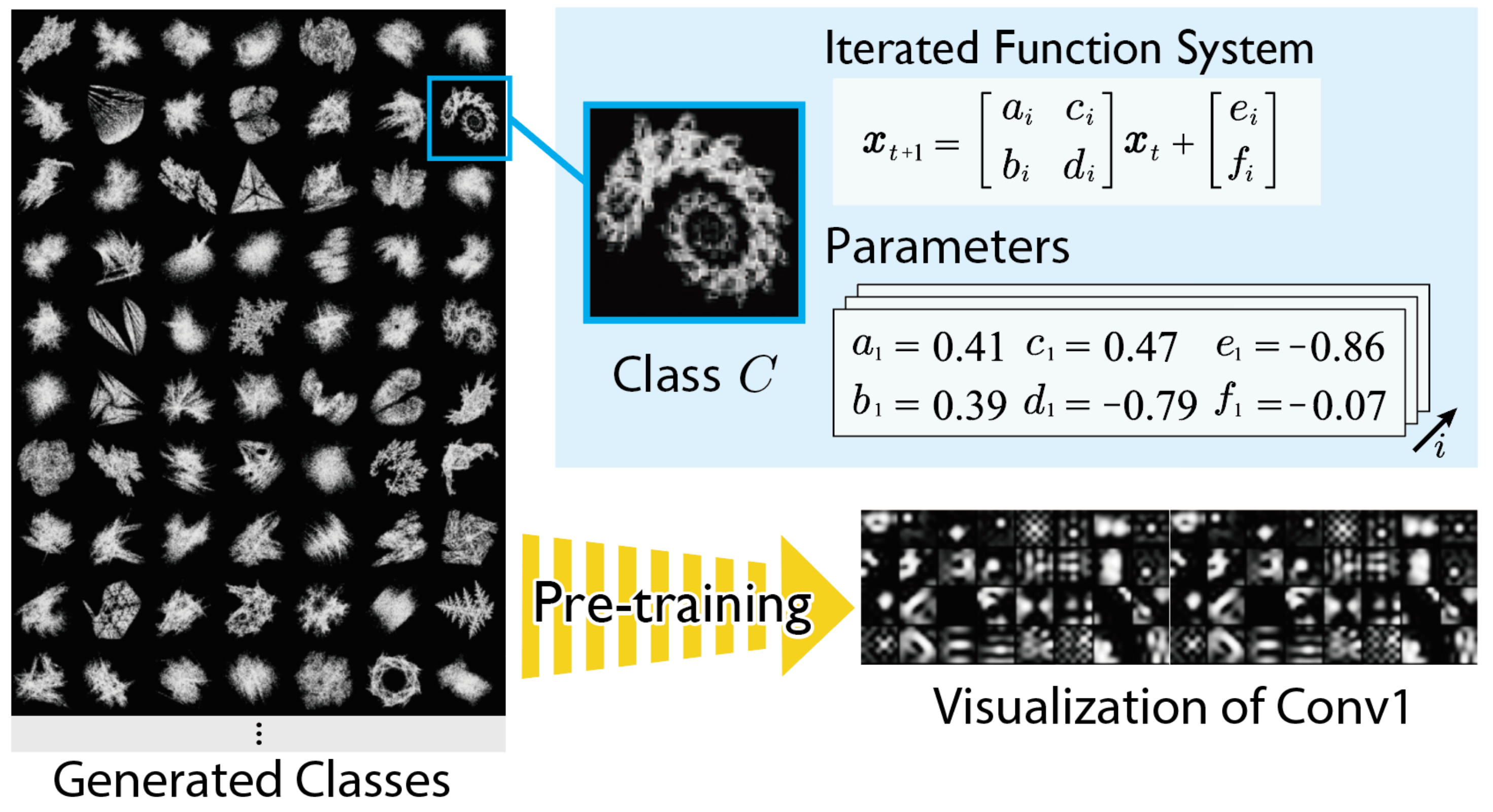}\label{fig:fractaldb}}
    \hspace{2pt}
  \subfigure[Accuracy transition among ImageNet-1k, FractalDB-1k and training from scratch.]{\includegraphics[width=0.42\linewidth]{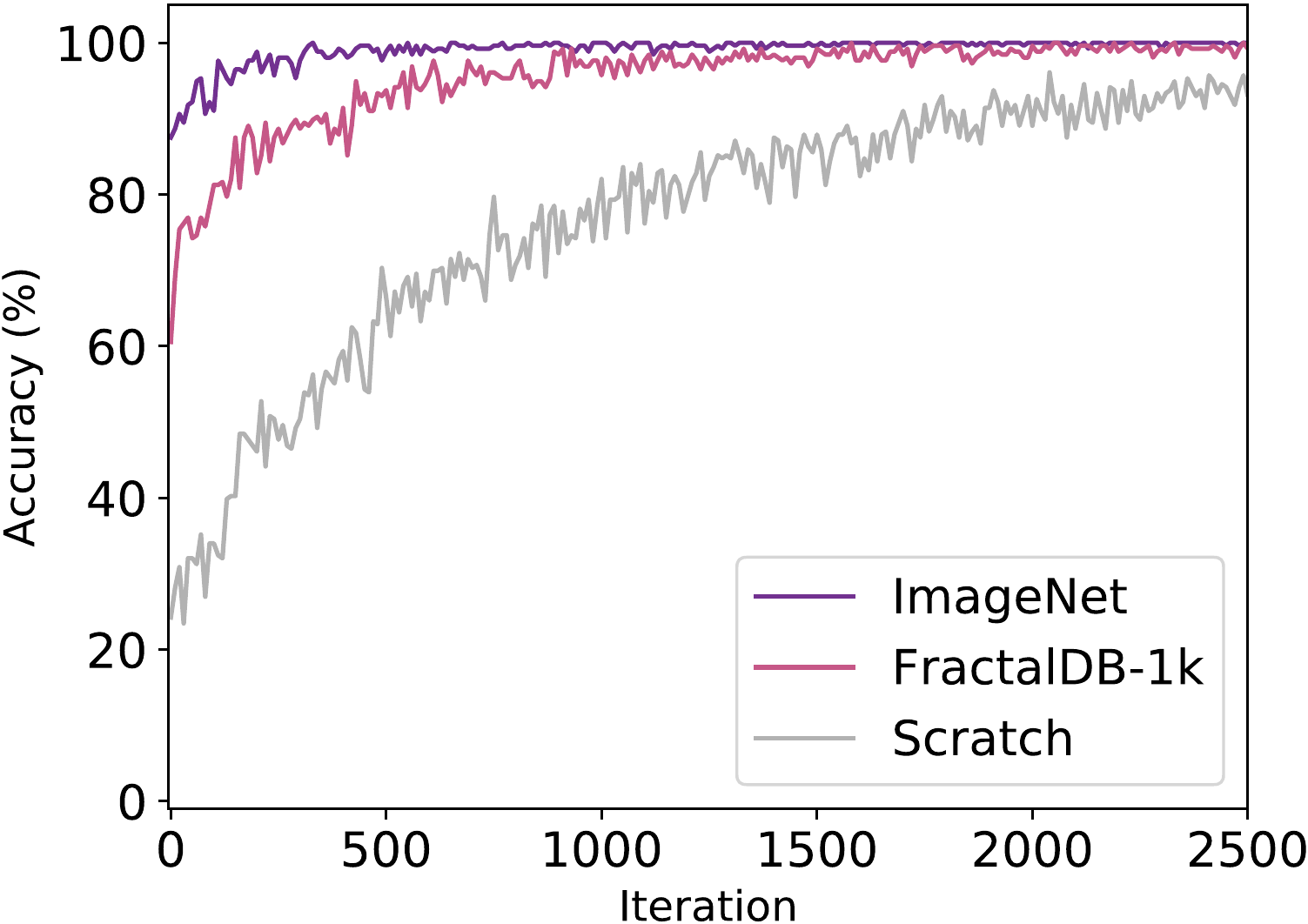}\label{fig:acc_compare}}
  \vspace{-0pt}
\caption{Proposed \textit{pre-training without natural images}  based on fractals, which is a natural formula existing in the real world (Formula-driven Supervised Learning). We automatically generate a large-scale labeled image dataset based on an iterated function system (IFS).}
\label{fig:loss_acc_comp}
\end{figure*}

\section{Introduction}

The introduction of sophisticated pre-training image representation has lead to a great expansion of the potential of image recognition. Image representations with e.g., the ImageNet/Places pre-trained convolutional neural networks (CNN), has without doubt become the most important breakthrough in recent years~\cite{imagenet,places}. We had lots to learn from the ImageNet project, such as huge amount of annotations done by crowdsourcing and well-organized categorization based on WordNet~\cite{wordnet}. However, due to the fact that the annotation was done by a large number of unspecified people, most of whom are unknowledgeable and not experts in image classification and the corresponding areas, the dataset contains mistaken, privacy-violated, and ethics-related labels~\cite{BuolamwiniMLR2018,YangFAT2020}. This limits the ImageNet to only non-commercial usage because the images included in the dataset does not clear the right related issues. We believe that this aspect of pre-trained models significantly narrows down the prospects of vision-based recognition.

We begin by considering what a pre-trained CNN model with a million natural images is. In most cases, representative image datasets consist of natural images taken by a camera that express a projection of the real world. Although the space of image representation is enormous, a CNN model has been shown to be capable of recognition of natural images from among around one million natural images from the ImageNet dataset. We believe that labeled images on the order of millions have a great potential to improve image representation as a pre-trained model. However, at the moment, a curious question occurs: \textit{Can we accomplish pre-training without any natural images for parameter fine-tuning on a dataset including natural images?} To the best of our knowledge, the ImageNet/Places pre-trained models have not been replaced by a model trained without natural images.
Here, we deeply consider pre-training without natural images. In order to replace the models pre-trained with natural images, we attempt to find a method for automatically generating images. Automatically generating a large-scale labeled image dataset is challenging, however, a model pre-trained without natural images makes it possible to solve problems related to privacy, copyright, and ethics, as well as issues related to the cost of image collection and labeling.

Unlike a synthetic image dataset, could we automatically make image patterns and their labels with image projection from a mathematical formula? 
Regarding synthetic datasets, the SURREAL dataset~\cite{VarolCVPR2017} has successfully made training samples of estimating human poses with human-based motion capture (mocap) and background. 
In contrast, our Formula-driven Supervised Learning and the generated formula-driven image dataset has a great potential to automatically generate an image pattern and a label. For example, we consider using \textit{fractals}, a sophisticated natural formula~\cite{Mandelbrot1983}. 
Generated fractals can differ drastically with a slight change in the parameters, and can often be distinguished in the real-world.
Most natural objects appear to be composed of complex patterns, but fractals allow us to understand and reproduce these patterns.

We believe that the concept of pre-training without natural images can simplify large-scale DB construction with formula-driven image projection in order to efficiently use a pre-trained model. Therefore, the formula-driven image dataset that includes automatically generated image patterns and labels helps to efficiently solve some of the current issues involved in using a CNN, namely, large-scale image database construction without human annotation and image downloading. 
Basically, the dataset construction does not rely on any natural images (e.g. ImageNet~\cite{imagenet} or Places~\cite{places}) or closely resembling synthetic images (e.g., SURREAL~\cite{VarolCVPR2017}). The present paper makes the following contributions.

The concept of pre-training without natural images provides a method by which to automatically generate a large-scale image dataset complete with image patterns and their labels. In order to construct such a database, through exploration research, we experimentally disclose ways to automatically generate categories using fractals.
The present paper proposes two sets of randomly searched fractal databases generated in such a manner: FractalDB-1k/10k, which consists of 1,000/10,000 categories (see the supplementary material for all FractalDB-1k categories). See Figure~\ref{fig:fractaldb} for Formula-driven Supervised Learning from categories of FractalDB-1k. Regarding the proposed database, the FractalDB pre-trained model outperforms some models pre-trained by human annotated datasets (see Table~\ref{tab:comparison} for details). Furthermore, Figure~\ref{fig:acc_compare} shows that FractalDB pre-training accelerated the convergence speed, which was much better than training from scratch and similar to ImageNet pre-training.

\vspace{-0pt}\section{Related work}

\textbf{Pre-training on Large-scale Datasets.} A number of large-scale datasets have been released for exploring how to extract an image representation, e.g., image classification~\cite{imagenet,places}, object detection~\cite{pascalvoc,LinECCV2014,openimages}, and video classification~\cite{KayarXiv2017,MonfortTPAMI2019}. These datasets have contributed to improving the accuracy of DNNs when used as (pre-)training. Historically, in multiple aspects of evaluation, the ImageNet pre-trained model has been proved to be strong in transfer learning~\cite{DonahueICML2014,HuhNIPS2016WS,KornblithCVPR2019}. Moreover, several larger-scale datasets have been proposed, e.g., JFT-300M~\cite{jft300m} and IG-3.5B~\cite{ig3.5b}, for further improving the pre-training performance. 

We are simply motivated to find a method to automatically generate a pre-training dataset without any natural images for acquiring a learning representation on image datasets. We believe that the proposed concept of pre-training without natural images will surpass the methods mentioned above in terms of fairness, privacy-violated, and ethics-related labels, in addition to the burdens of human annotation and image download.

\textbf{Learning Frameworks.} Supervised learning with well-studied architectures is currently the most promising framework for obtaining strong image representations~\cite{alexnet,vggnet,inception,resnet,resnext,HowardarXiv2017,SandlerarXiv2018,HowardICCV2019}.
Recently, the research community has been considering how to decrease the volume of labeled data with \{un, weak, self\}-supervised learning in order to avoid human labeling.
In particular, self-supervised learning can be used to create a pre-trained model in a cost-efficient manner by using {\it obvious} labels.
The idea is to make a simple but suitable task, called a pre-text task~\cite{DoerschICCV2015,NorooziECCV2016,NorooziCVPR2018,ZhangECCV2016,NorooziICCV2017,GidarisICLR2018}. 
Though the early approaches (e.g., jigsaw puzzle~\cite{NorooziECCV2016}, image rotation~\cite{GidarisICLR2018}, and colorization~\cite{ZhangECCV2016}) were far from an alternative to human annotation, the more recent approaches (e.g., DeepCluster~\cite{deepcluster}, MoCo~\cite{HeCVPR2020}, and SimCLR~\cite{ChenICML2020}) are becoming closer to a human-based supervision like ImageNet.

The proposed framework is complementary to these studies because the above learning frameworks focus on how to represent a natural image based on an existing dataset. Unlike these studies, the proposed framework enables the generation of new image patterns based on a mathematical formula in addition to training labels. The SSL framework can replace the manual labeling supervised by human knowledge, however, there still exists the burdens of image downloading, privacy violations and unfair outputs.

\textbf{Mathematical formula for image projection.} One of the best-known formula-driven image projections is fractals. Fractal theory has been discussed in a long period 
(e.g.,~\cite{Mandelbrot1983,LandiniIOVS95,SmithJNM96}). Fractal theory has been applied to rendering a graphical pattern in a simple equation~\cite{fractals_everywhere,fractals_algorithms,3dfractal} and constructing visual recognition models~\cite{PentlandTPAMI84,VarmaICCV2007,XuIJCV09,LarssonICLR2017}. Although a rendered fractal pattern loses its infinite potential for representation by projection to a 2D-surface, a human can recognize the rendered fractal patterns as natural objects.

Since the success of these studies relies on the fractal geometry of naturally occurring phenomena~\cite{Mandelbrot1983,Falconer2004}, our assumption that fractals can assist learning image representations for recognizing natural scenes and objects is supported. Other methods, namely, the Bezier curve~\cite{beziercurve} and Perlin noise~\cite{PerlinToG2002}, have also been discussed in terms of computational rendering. We also implement and compare these methods in the experimental section (see Table~\ref{tab:bezie_perlin}).

\vspace{-0pt}\section{Automatically generated large-scale dataset}

\begin{figure*}[t]
\begin{center}
   \includegraphics[width=0.95\linewidth]{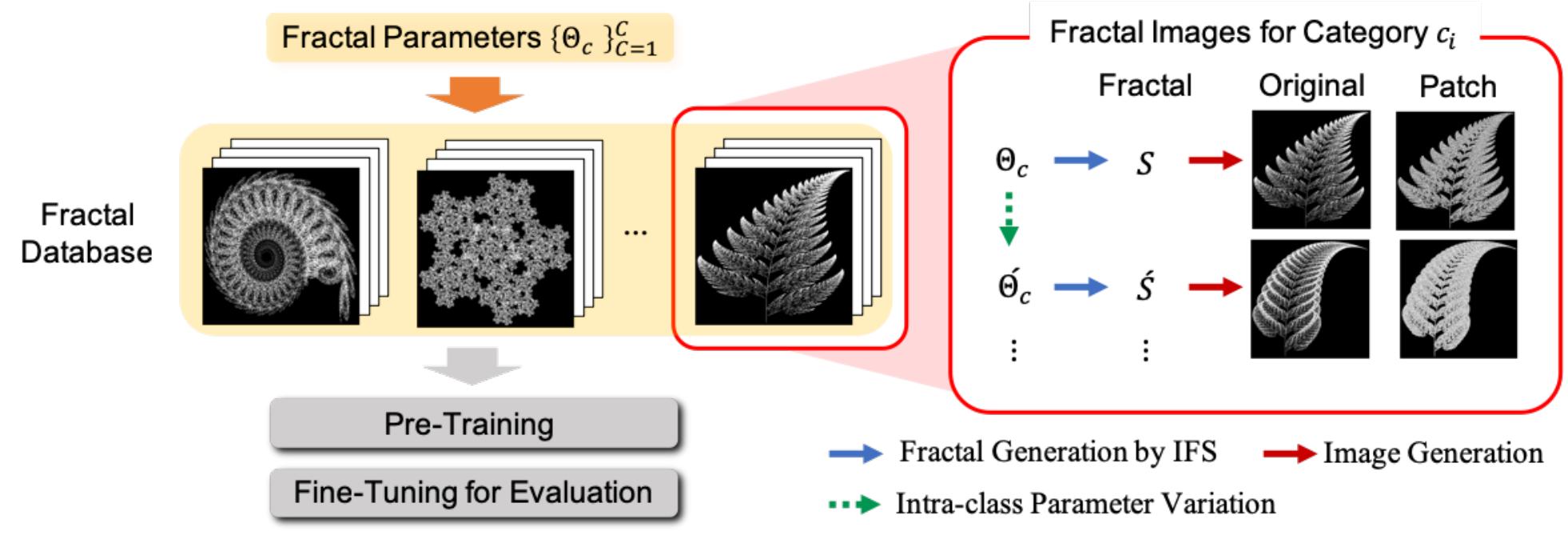}
   \vspace{-0pt}\caption{
   Overview of the proposed framework. Generating FractalDB: Pairs of an image $I_{j}$ and its fractal category $c_{j}$ are generated without human labeling and image downloading. Application to transfer learning: A FractalDB pre-trained convolutional network is assigned to conduct transfer learning for other datasets.}
\label{fig:concept}
\end{center}
\vspace{-0pt}
\end{figure*}

Figure~\ref{fig:concept} presents an overview of the Fractal DataBase (FractalDB), which consists of an infinite number of pairs of fractal images $I$ and their fractal categories $c$ with iterated function system (IFS)~\cite{fractals_everywhere}.
We chose fractal geometry because the function enables to render complex patterns with a simple equation that are closely related to natural objects.
All fractal categories are randomly searched (see Figure~\ref{fig:fractaldb}), and the intra-category instances are expansively generated by considering category configurations such as rotation and patch. (The augmentation is shown as $\theta \rightarrow \theta^{'}$ in Figure~\ref{fig:concept}.)
 
In order to make a pre-trained CNN model, the FractalDB is applied to each training of the parameter optimization as follows.
(i) Fractal images with paired labels are randomly sampled by a mini batch $B=\{(I_{j},c_{j})\}_{j=1}^{b}$. (ii) Calculate the gradient of $B$ to reduce the loss. (iii) Update the parameters. Note that we replace the pre-training step, such as the ImageNet pre-trained model. We also conduct the fine-tuning step as well as plain transfer learning (e.g., ImageNet pre-training and CIFAR-10 fine-tuning).

\subsection{Fractal image generation}
\label{sec:fractalgenerationsystems}
In order to construct fractals, we use IFS~\cite{fractals_everywhere}.
In fractal analysis, an IFS is defined on a complete metric space $\mathcal{X}$ by
\begin{align}
\mbox{IFS} = \{\mathcal{X}; w_{1},w_{2},\cdots,w_{N}; p_{1},p_{2},\cdots,p_{N}\},
\end{align}\vspace{-0pt}
where $w_{i}:\mathcal{X} \rightarrow \mathcal{X}$ are transformation functions,
$p_{i}$ are probabilities having the sum of 1,
and $N$ is the number of transformations.

Using the IFS, a fractal $S = \{\bm{x}_{t}\}_{t=0}^{\infty} \in \mathcal{X}$ is constructed by the random iteration algorithm~\cite{fractals_everywhere}, which repeats the following two steps for $t=0,1,2,\cdots$ from an initial point $\bm{x}_{0}$. (i) Select a transformation $w^{*}$ from $\{w_{1},\cdots,w_{N}\}$ with pre-defined probabilities $p_{i} = p(w^{*}=w_{i})$ to determine the $i$-th transformation. (ii) Produce a new point $\bm{x}_{t+1} = w^{*}(\bm{x}_{t})$.

Since the focus herein is on representation learning for image recognition,
we construct fractals in the 2D Euclidean space $\mathcal{X} = \mathbb{R}^2$.
In this case, each transformation is assumed in practice to be an affine transformation ~\cite{fractals_everywhere}, which has a set of six parameters $\theta_{i} = (a_{i},b_{i},c_{i},d_{i}, e_{i}, f_{i})$ for rotation and shifting:
\begin{align}
 w_{i}(\bm{x};\theta_{i}) = 
 \begin{bmatrix}
 a_{i} & b_{i} \\
 c_{i} & d_{i} \\
 \end{bmatrix}
 \bm{x} +
 \begin{bmatrix}
 e_{i} \\
 f_{i} \\
 \end{bmatrix}.
\end{align}
An image representation of the fractal $S$ is obtained by drawing dots on a black background. The details of this step with its adaptable parameters is explained in Section~\ref{sec:categorydefinition}.

\subsection{Fractal categories}
\label{sec:fractalcategories}

Undoubtedly, automatically generating categories for pre-training of image classification is a challenging task. Here, we associate the categories with fractal parameters $a$--$f$. As shown in the experimental section, we successfully generate a number of pre-trained categories on FractalDB (see Figure~\ref{fig:acc_labelnoise}) through formula-driven image projection by an IFS.

Since an IFS is characterized by a set of parameters and their corresponding probabilities, i.e., $\Theta = \{(\theta_{i},p_{i})\}_{i=1}^{N}$, we assume that a fractal category has a fixed $\Theta$ and propose 1,000 or 10,000 randomly searched fractal categories (FractalDB-1k/10k). The reason for 1,000 categories is closely related to the experimental result for various \#categories in Figure~\ref{fig:cifar_catins}.

{\bf FractalDB-1k/10k} consists of 1,000/10,000 different fractals (examples shown in Figure~\ref{fig:fractaldb}),
the parameters of which are automatically generated by repeating the following procedure.
First, $N$ is sampled from a discrete uniform distribution, $\mathbb{N} = \{2,3,4,5,6,7,8\}$.
Second, the parameter $\theta_{i}$ for the affine transformation is sampled from the uniform distribution on $[-1,1]^{6}$ for $i = 1,2,\cdots,N$.
Third, $p_{i}$ is set to
$p_{i} = (\det A_{i}) /$ $(\sum_{i=1}^{N} \det A_{i})$ where $A_{i} = (a_{i},b_{i};c_{i},d_{i})$ is a rotation matrix of the affine transformation.
Finally, $\Theta_{i} = \{(\theta_{i},p_{i})\}_{i=1}^{N}$ is accepted as a new category if the filling rate $r$ of the representative image of its fractal $S$ is investigated in the experiment (see Table~\ref{tab:fillingrate}). The filling rate $r$ is calculated as the number of pixels of the fractal with respect to the total number of pixels of the image.

\begin{figure}[t]
\begin{center}
   \includegraphics[width=0.95\linewidth]{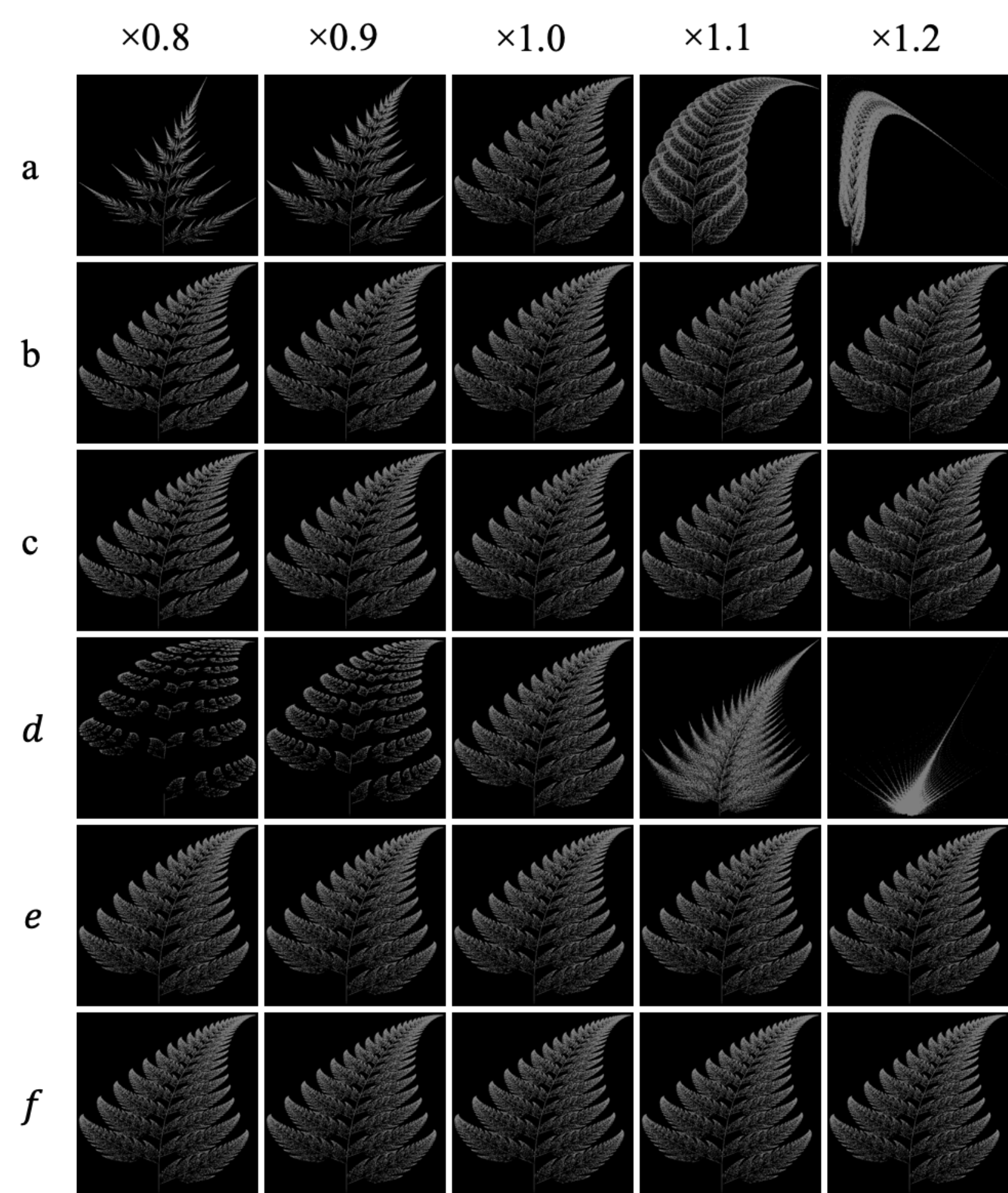}
   \caption{Intra-category augmentation of a {\it leaf} fractal. Here, $a_{i}$, $b_{i}$, $c_{i}$, and $d_{i}$ are for rotation, and $e_{i}$ and $f_{i}$ are for shifting.}
\label{fig:fractalrs1k_aug}
\end{center}
\vspace{-0pt}
\end{figure}

\subsection{Adaptable parameters for FractalDB}
\label{sec:categorydefinition}

As described in the experimental section, we investigated the several parameters related to fractal parameters and image rendering. The types of parameters are listed as follows.

\textbf{\#category and \#instance.} We believe that the effects of \#instance on intra-category are the most effective in the pre-training task. First, we change the two parameters from 16 to 1,000 as \{16, 32, 64, 128, 256, 512, 1,000\}. 

\textbf{Patch vs. Point.} We apply a 3$\times$3 patch filter to generate fractal images in addition to the rendering at each 1$\times$1 point. The patch filter makes variation in the pre-training phase. We repeat the following process $t$ times. We set a pixel $(u,v)$, and then a random dot(s) with a 3$\times$3 patch is inserted in the sampled area.

\textbf{Filling rate $r$.} We set the filling rate from 0.05 (5\%) to 0.25 (25\% at 5\% intervals, namely, \{0.05, 0.10, 0.15, 0.20,  0.25\}. Note that we could not get any randomized category at a filling rate of over 30\%.

\textbf{Weight of intra-category fractals ($w$).} In order to generate an intra-category image, the parameters for an image representation are varied.
Intra-category images are generated by changing one of the parameters $a_{i},b_{i},c_{i},d_{i}$, and $e_{i},f_{i}$ with weighting parameter $w$. The basic parameter is from $\times$0.8 to $\times$1.2 at intervals of 0.1, i.e., \{0.8, 0.9, 1.0, 1.1,  1.2\}).
Figure~\ref{fig:fractalrs1k_aug} shows an example of the intra-category variation in fractal images. We believe that various intra-category images help to improve the representation for image recognition.

\textbf{\#Dot ($t$) and image size ($W$, $H$).} We vary the parameters $t$ as \{100K, 200K, 400K, 800K\} and ($W$ and $H$) as \{256, 362, 512, 764, 1024\}. The averaged parameter fixed as the grayscale means that the pixel value is ($r$, $g$, $b$) = (127, 127, 127) (in the case in which the pixel values are 0 to 255).

\section{Experiments}

In a set of experiments, we investigated the effectiveness of FractalDB and how to construct categories with the effects of configuration, as mentioned in Section~\ref{sec:categorydefinition}.
We then quantitatively evaluated and compared the proposed framework with Supervised Learning (ImageNet-1k and Places365, namely ImageNet~\cite{imagenet} and Places~\cite{places} pre-trained models) and Self-supervised Learning (Deep Cluster-10k~\cite{deepcluster}) on several datasets~\cite{cifar,imagenet,places,pascalvoc,omniglot}.

\begin{figure*}[t]
\centering
\subfigure[CIFAR10]{\includegraphics[width=0.23\linewidth]{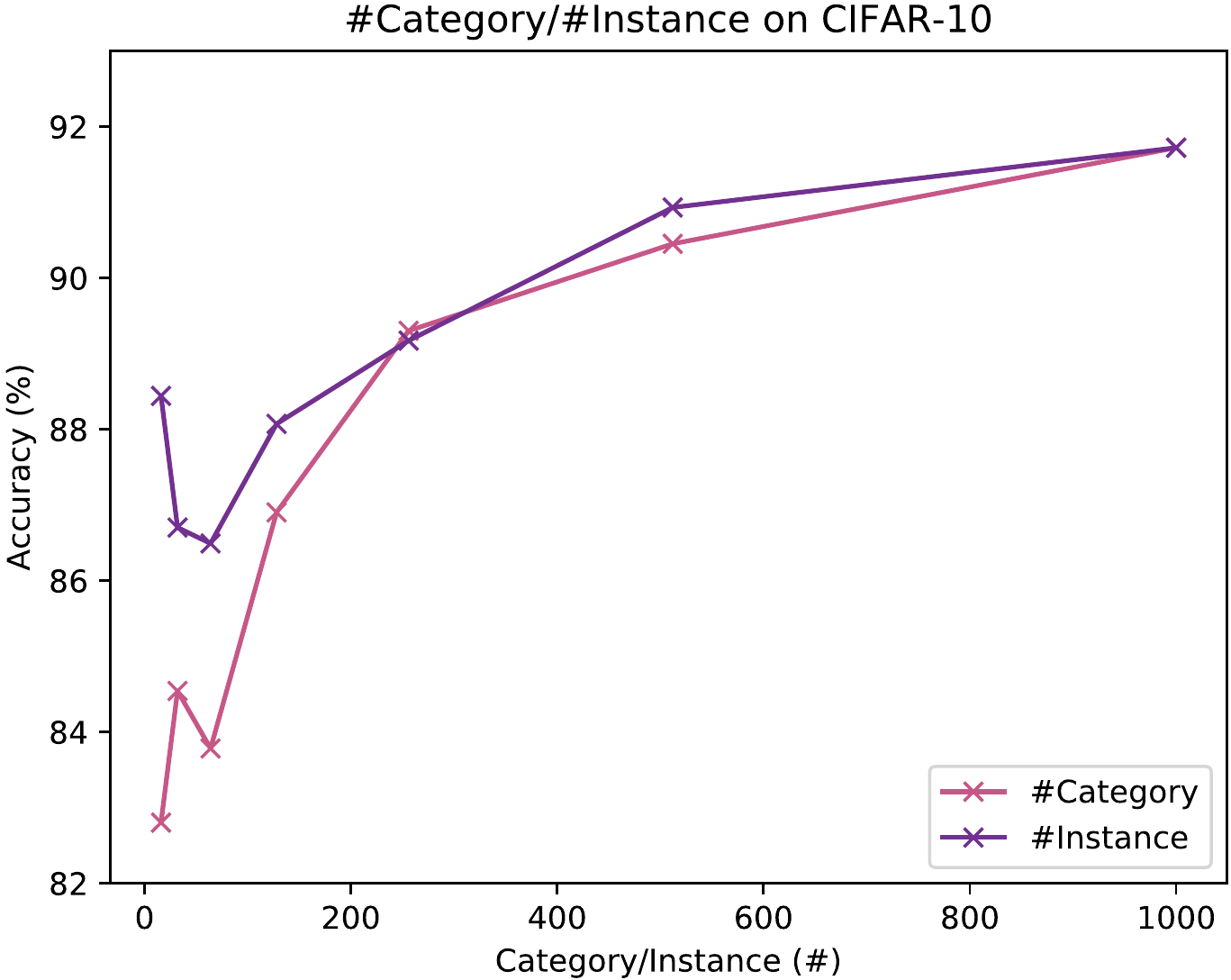}
\label{fig:cifar10_catins}}
\subfigure[CIFAR100]{\includegraphics[width=0.23\linewidth]{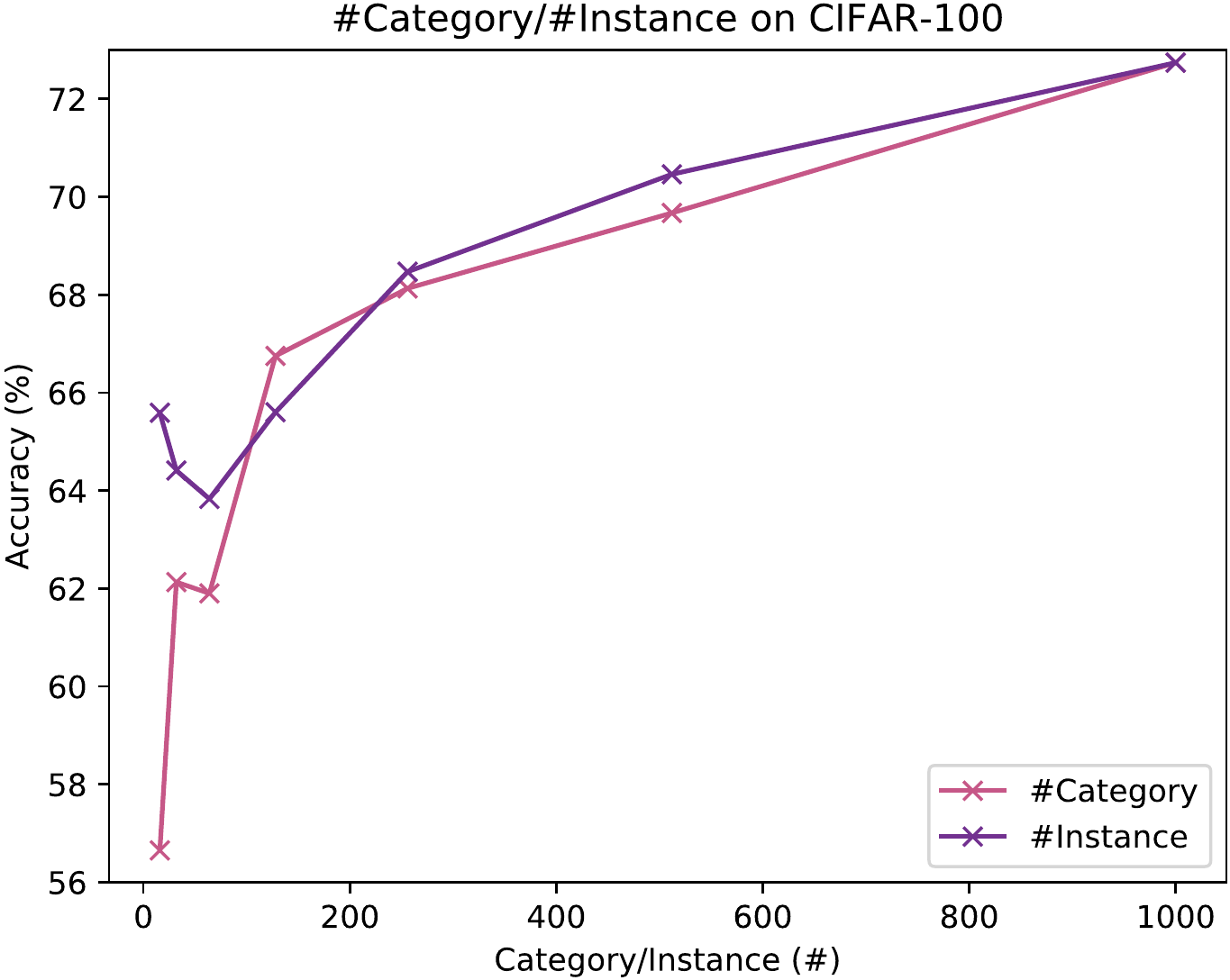}
\label{fig:cifar100_catins}}
\subfigure[ImageNet100]{\includegraphics[width=0.23\linewidth]{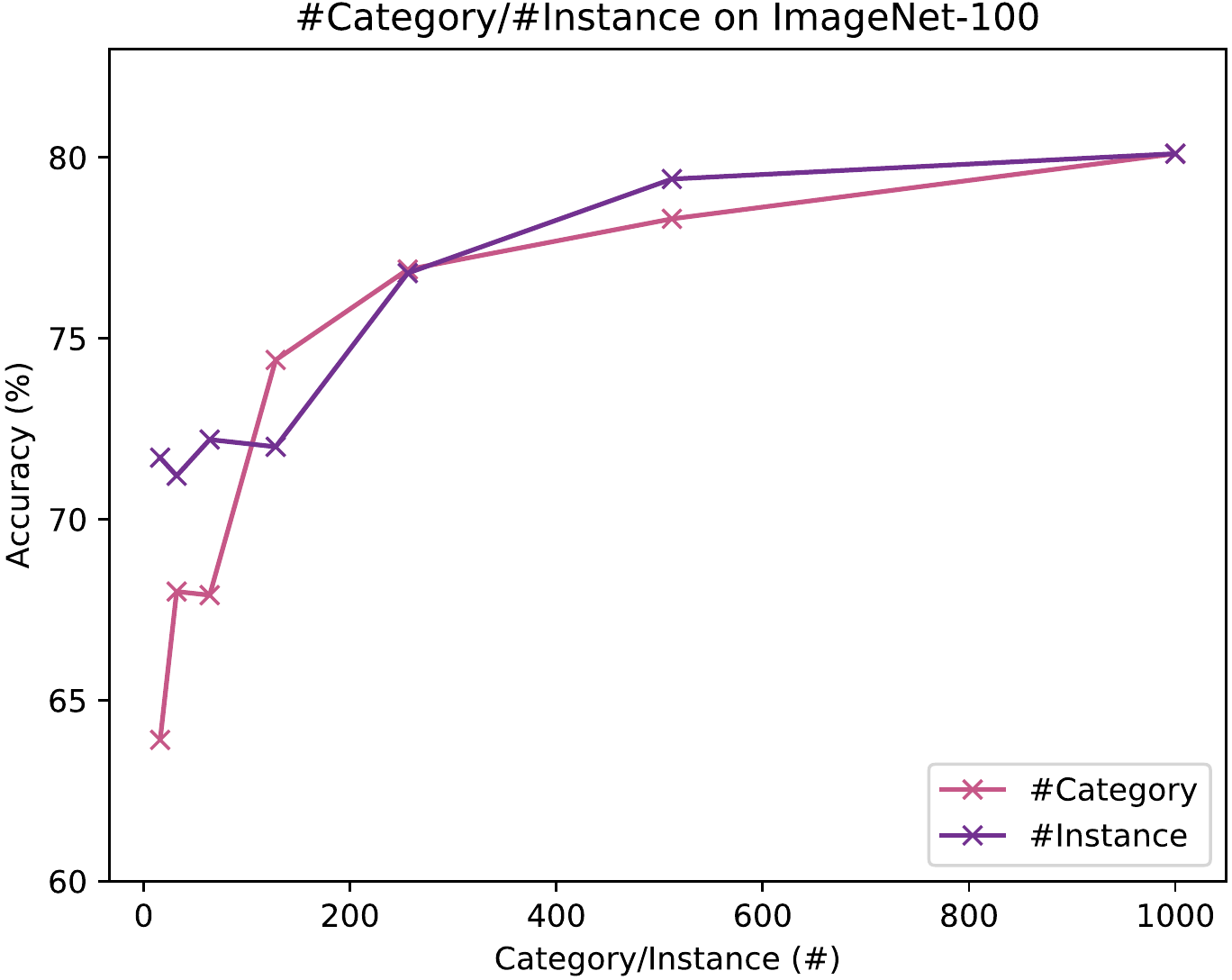}
\label{fig:imagenet100_catins}}
\subfigure[Places30]{\includegraphics[width=0.23\linewidth]{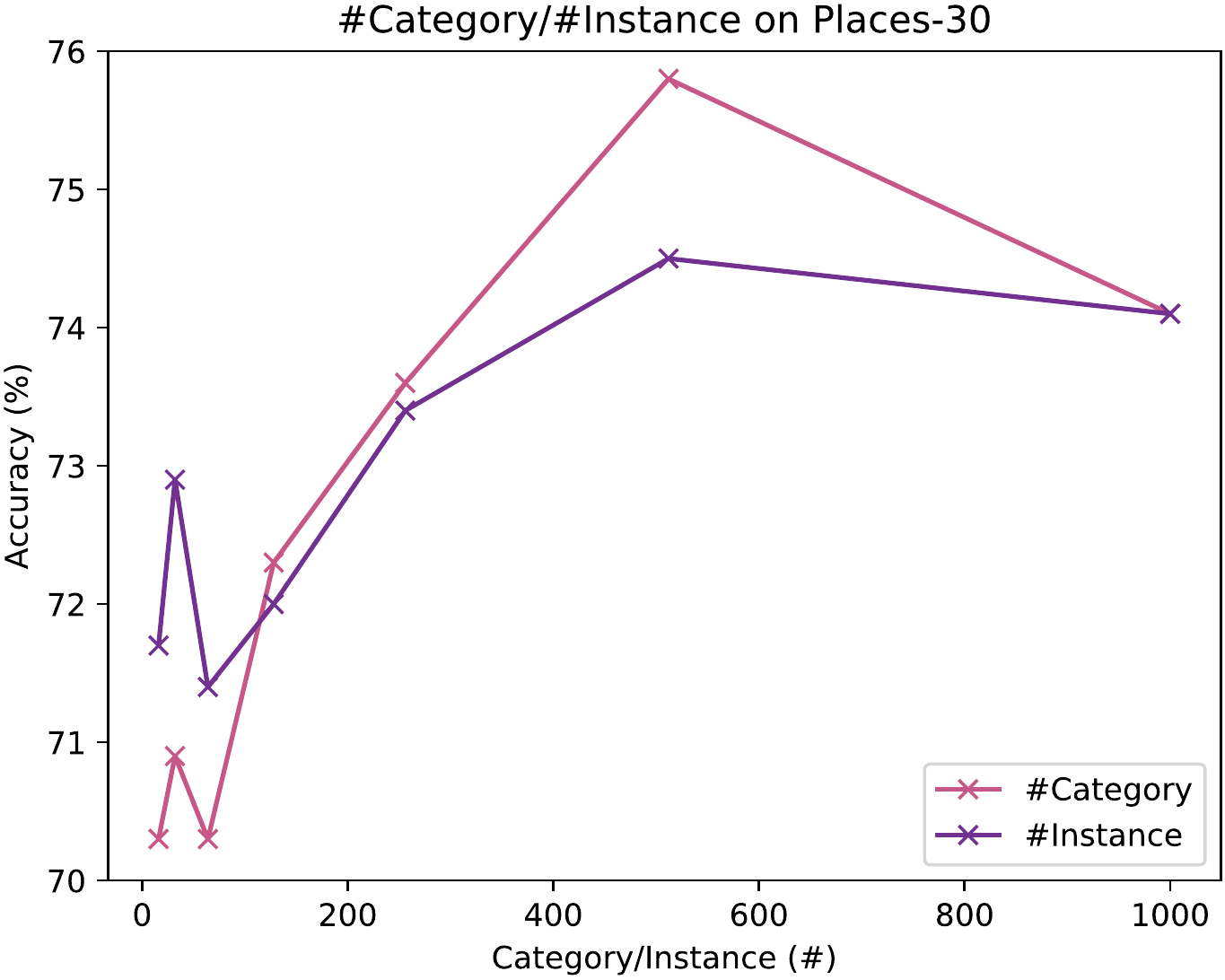}
\label{fig:places30_catins}}
\caption{Effects of \#category and \#instance on the CIFAR-10/100, ImageNet100 and Places30 datasets. The other parameter is fixed at 1,000, e.g. \#Category is fixed at 1,000 when \#Instance changed by \{16, 32, 64, 128, 256, 512, 1,000\}.}
\label{fig:cifar_catins}
\vspace{-0pt}
\end{figure*}

In order to confirm the properties of FractalDB and compare our pre-trained feature with previous studies, we used the ResNet-50.
We simply replaced the pre-trained phase with our FractalDB (e.g., FractalDB-1k/10k), without changing the fine-tuning step. Moreover, in the usage of fine-tuning datasets, we conducted a standard training/validation.
Through pre-training and fine-tuning, we assigned the momentum stochastic gradient descent (SGD)~\cite{Bottou2010} with a value 0.9, a basic batch size of 256, and initial values of the learning rate of 0.01. The learning rate was multiplied by 0.1 when the learning epoch reached 30 and 60. Training was performed up to epoch 90. Moreover, the input image size was cropped by  $224\times224$ [pixel] from a $256\times256$ [pixel] input image.

\subsection{Exploration study}
\label{sec:preliminarystudy}

In this subsection, we explored the configuration of formula-driven image datasets regarding Fractal generation by using CIFAR-10/100 (C10, C100), ImageNet-100 (IN100), and Places-30 datasets (P30) datasets (see the supplementary material for category lists in ImageNet-100 and Places-30). The parameters corresponding to those mentioned in Section~\ref{sec:categorydefinition}.

\textbf{\#category and \#instance} (see Figures~\ref{fig:cifar10_catins}, \ref{fig:cifar100_catins}, \ref{fig:imagenet100_catins} and \ref{fig:places30_catins})$\rightarrow$ Here, the larger values tend to be better. Figure~\ref{fig:cifar_catins} indicates the effects of category and instance. We investigated the parameters with \{16, 32, 64, 128, 256, 512, 1,000\} on both properties. At the beginning, a larger parameter in pre-training tends to improve the accuracy in fine-tuning on all the datasets. With C10/100, we can see +7.9/+16.0 increases on the performance rate from 16 to 1,000 in \#category. The improvement can be confirmed, but is relatively small for the \#instance per category. The rates are +5.2/+8.9 on C10/100.

Hereafter, we assigned 1,000 [category] $\times$ 1,000 [instance] as a basic dataset size and tried to train 10k categories since the \#category parameter is more effective in improving the performance rates.

\textbf{Patch vs. point} (see Table~\ref{tab:patchvspoint})$\rightarrow$ Patch with 3 $\times$ 3 [pixel] is better. Table~\ref{tab:patchvspoint} shows the difference between $3 \times 3$ patch rendering and $1 \times 1$ point rendering. We can confirm that the $3 \times 3$ patch rendering is better for pre-training with 92.1 vs. 87.4 (+4.7) on C10 and 72.0 vs. 66.1 (+5.9) on C100. Moreover, when comparing random patch pattern at each patch (random) to fixed patch in image rendering (fix), performance rates increased by \{+0.8, +1.6, +1.1, +1.8\} on \{C10, C100, IN100, P30\}.

\begin{table*}[t]
  \begin{minipage}[t]{.50\textwidth}
    \caption{Exploration: Patch vs. point.}
    \vspace{-0pt}
    \begin{center}
      \begin{tabular}{l|cccc} \hline
         & C10 & C100 & IN100 & P30 \\
        \hline \hline
        Point & 87.4 & 66.1 & 73.9 & 73.0 \\
        Patch (random) & \textbf{92.1} & \textbf{72.0} & \textbf{78.9} & \textbf{73.2} \\
        Patch (fix) & \textbf{92.9} & \textbf{73.6} & \textbf{80.0} & \textbf{75.0} \\
        \hline
      \end{tabular}
    \end{center}
    \label{tab:patchvspoint}
  \end{minipage}
  \hfill
  \begin{minipage}[t]{.50\textwidth}
    \caption{Exploration: Filling rate.}
    \begin{center}
      \begin{tabular}{l|cccc} \hline
         & C10 & C100 & IN100 & P30 \\
        \hline \hline
        .05 & 91.8 & \textbf{72.4} & 80.2 & 74.6 \\
        .10 & \textbf{92.0} & 72.3 & \textbf{80.5} & \textbf{75.5} \\
        .15 & 91.7 & 71.6 & 80.2 & 74.3 \\
        .20 & 91.3 & 70.8 & 78.8 & 74.7 \\
        .25 & 91.1 & 63.2 & 72.4 & 74.1 \\
        \hline
      \end{tabular}
    \end{center}
    \label{tab:fillingrate}
  \end{minipage}
  %
  \begin{minipage}[t]{.3\textwidth}
    \caption{Exploration: Weights}
    \vspace{-0pt}
    \begin{center}
      \begin{tabular}{l|cccc} \hline
         & C10 & C100 & IN100 & P30 \\
        \hline \hline
        .1 & 92.1 & 72.0 & 78.9 & 73.2 \\
        .2 & 92.4 & 72.7 & 79.2 & 73.9 \\
        .3 & 92.4 & 72.6 & 79.2 & 74.3 \\
        .4 & \textbf{92.7} & \textbf{73.1} & \textbf{79.6} & \textbf{74.9} \\
        .5 & 91.8 & 72.1 & 78.9 & 73.5 \\
        \hline
      \end{tabular}
    \end{center}
    \label{tab:weights}
  \end{minipage}
  \begin{minipage}[t]{.33\textwidth}
    \caption{Exploration: \#Dot.}
    \begin{center}
      \begin{tabular}{l|cccc} \hline
         & C10 & C100 & IN100 & P30 \\
        \hline \hline
        100k & \textbf{91.3} & 70.8 & 78.8 & 74.7 \\
        200k & 90.9 & \textbf{71.0} & 79.2 & \textbf{74.8} \\
        400k & 90.4 & 70.3 & \textbf{80.0} & 74.5 \\
        \hline
      \end{tabular}
    \end{center}
    \label{tab:numofdot}
  \end{minipage}
  \begin{minipage}[t]{.3\textwidth}
    \caption{Exploration: Image size}
    \begin{center}
      \begin{tabular}{l|cccc} \hline
         & C10 & C100 & IN100 & P30 \\
        \hline \hline
        256 & \textbf{92.9} & \textbf{73.6} & 80.0 & 75.0 \\
        362 & 92.2 & 73.2 & \textbf{80.5} & \textbf{75.1} \\
        512 & 90.9 & 71.0 & 79.2 & 73.0 \\
        724 & 90.8 & 71.0 & 79.2 & 73.0 \\
        1024 & 89.6 & 68.6 & 77.5 & 71.9 \\
        \hline
      \end{tabular}
    \end{center}
    \label{tab:imagesize}
  \end{minipage}
\end{table*}

\textbf{Filling rate} (see Table~\ref{tab:fillingrate})$\rightarrow$ 0.10 is better, but there is no significant change with \{0.05, 0.10, 0.15\}. The top scores for each dataset and the parameter are 92.0, 80.5 and 75.5 with a filling rate of 0.10 on C10, IN100 and P30, respectively. Based on these results, a filling rate of 0.10 appears to be better. 

\textbf{Weight of intra-category fractals} (see Table~\ref{tab:weights})$\rightarrow$ Interval 0.4 is the best. A larger variance of intra-category tends to perform better in pre-training. Starting from the basic parameter at intervals of 0.1 with \{0.8, 0.9, 1.0, 1.1,  1.2\} (see Figure~\ref{fig:fractalrs1k_aug}), we varied the intervals as 0.1, 0.2, 0.3, 0.4, and 0.5. For the case in which the interval is 0.5, we set \{0.01, 0.5, 1.0, 1.5, 2.0\} in order to avoid the weighting value being set as zero. A higher variance of intra-category tends to provide higher accuracy. We confirm that the accuracies varied as \{92.1, 92.4, 92.4, \textbf{92.7}, 91.8\} on C10, where 0.4 is the highest performance rate (92.7), but 0.5 decreases the recognition rate (91.8). We used the weight value with a 0.4 interval, i.e., \{0.2, 0.6, 1.0, 1.4, 1.8\}.

\textbf{\#Dot} (see Table~\ref{tab:numofdot})$\rightarrow$ We selected 200k by considering the accuracy and rendering time. The best parameters for each configurations are 100K on C10 (91.3), 200k on C100/P30 (71.0/74.8) and 400k on IN100 (80.0). Although a larger value is suitable on IN100, a lower value tends to be better on C10, C100, and P30. For the \#dot parameter, 200k is the most balanced in terms of rendering speed and accuracy.

\textbf{Image size} (see Table~\ref{tab:imagesize})$\rightarrow$ 256 $\times$ 256 or 362 $\times$ 362 is better. In terms of image size, $256 \times 256$ [pixel] and $362 \times 362$ [pixel] have similar performances, e.g., 73.6 (256) vs. 73.2 (362) on C100. A larger size, such as $1{,}024 \times 1{,}024$, is sparse in the image plane. Therefore, the fractal image projection produces better results in the cases of $256 \times 256$ [pixel] and $362 \times 362$ [pixel].


Moreover, we have additionally conducted two configurations with grayscale and color FractalDB. However, the effect of the color property appears not to be strong in the pre-training phase.

\subsection{Comparison to other pre-trained datasets}

We compared \textbf{Scratch} from random parameters, \textbf{Places-30/365}~\cite{places}, \textbf{ImageNet-100/1k} (ILSVRC'12)~\cite{imagenet}, and \textbf{FractalDB-1k/10k} in Table~\ref{tab:comparison}. 
Since our implementation is not completely the same as a representative learning configuration, we implemented the framework fairly with the same parameters and compared the proposed method (FractalDB-1k/10k) with a baseline (Scratch, DeepCluster-10k, Places-30/365, and ImageNet-100/1k).

The proposed FractalDB pre-trained model recorded several good performance rates. We respectively describe them by comparing our Formula-driven Supervised Learning with Scratch, Self-supervised and Supervised Learning.

\textbf{Comparison to training from scratch.} FractalDB-1k / 10k pre-trained models recorded much higher accuracies than models trained from scratch on relatively small-scale datasets (C10/100, VOC12 and OG). In case of fine-tuning on large-scale datasets (ImageNet/Places365), the effect of pre-training was relatively small. However, in fine-tuning on Places 365, the FractalDB-10k pre-trained model helped to improve the performance rate which was also higher than ImageNet-1k pre-training (FractalDB-10k 50.8 vs. ImageNet-1k 50.3).

\begin{table*}[t]
\begin{center}
\caption{Classification accuracies of the Ours (FractalDB-1k/10k), Scratch, DeepCluster-10k (DC-10k), ImageNet-100/1k and Places-30/365 pre-trained models on representative pre-training datasets. We show the types of pre-trained image (Pre-train Img; which includes \{Natural Image (Natural), Formula-driven Image (Formula)\}) and Supervision types (Type; which includes \{Self-supervision, Supervision, Formula-supervision\}). We employed CIFAR-10 (C10), CIFAR-100 (C100), ImageNet-1k (IN1k), Places-365 (P365), classfication set of Pascal VOC 2012 (VOC12) and Omniglot (OG) datasets. The \underline{\textbf{bold and underlined}} values show the best scores, and \textbf{bold} values indicate the second best scores.}
\begin{tabular}{lcc|cccccc} \hline
 Method & Pre-train Img & Type & C10 & C100 & IN1k & P365 & VOC12 & OG \\ \hline\hline
 Scratch & -- & -- & 87.6 & 62.7 & \underline{\textbf{76.1}} & 49.9 & 58.9 & 1.1 \\
 DC-10k & Natural & Self-supervision & 89.9 & 66.9 & 66.2 & \underline{\textbf{51.5}} & 67.5 & 15.2 \\ 
 Places-30 & Natural & Supervision & 90.1 & 67.8 & 69.1 & -- & 69.5 & 6.4 \\ 
 Places-365 & Natural & Supervision & \textbf{94.2} & 76.9 & 71.4 & -- & \textbf{78.6} & 10.5 \\ 
 ImageNet-100 & Natural & Supervision & 91.3 & 70.6 & -- & 49.7 & 72.0 & 12.3 \\ 
 ImageNet-1k & Natural & Supervision & \underline{\textbf{96.8}} & \underline{\textbf{84.6}} & -- & 50.3 & \underline{\textbf{85.8}} & 17.5\\ \hline
 \rowcolor{gray!20}FractalDB-1k & Formula & Formula-supervision & 93.4 & 75.7 & 70.3 & 49.5 & 58.9 & \textbf{20.9} \\
 \rowcolor{gray!20}FractalDB-10k & Formula & Formula-supervision & 94.1 & \textbf{77.3} & \textbf{71.5} & \textbf{50.8} & 73.6 & \underline{\textbf{29.2}} \\
  \hline
\end{tabular}
\label{tab:comparison}
\end{center}
\vspace{-0pt}
\end{table*}

\textbf{Comparison to Self-supervised Learning.} We assigned the DeepCluster-10k~\cite{deepcluster} to compare the automatically generated image categories. The 10k indicates the pre-training with 10k categories. We believe that the auto-annotation with DeepCluster is the most similar method to our formula-driven image dataset. The DeepCluster-10k also assigns the same category to images that has similar image patterns based on K-means clustering. Our FractalDB-1k/10k pre-trained models outperformed the DeepCluster-10k on five different datasets, e.g., FractalDB-10k 94.1 vs. DeepCluster 89.9 (C10), 77.3 vs. DeepCluster-10k 66.9 (C100). Our method is better than the DeepCluster-10k which is a self-supervised learning method to train a feature representation in image recognition.

\textbf{Comparison to Supervised Learning.} We compared four types of supervised pre-training (e.g., ImageNet-1k and Places-365 datasets and their limited categories ImageNet-100 and Places-30 datasets). ImageNet-100 and Places-30 are subsets of ImageNet-1k and Places-365. The numbers correspond to the number of categories. At the beginning, our FractalDB-10k surpassed the ImageNet-100/Places-30 pre-trained models at all fine-tuning datasets. The results show that our framework is more effective than the pre-training with subsets from ImageNet-1k and Places365.

We compare the supervised pre-training methods which are the most promising pre-training approach ever. Although our FractalDB-1k/10k cannot beat them at all settings, our method partially outperformed the ImageNet-1k pre-trained model on Places-365 (FractalDB-10k 50.8 vs. ImageNet-1k 50.3) and Omniglot (FractalDB-10k 29.2 vs. ImageNet-1k 17.5) and Places-365 pre-trained model on CIFAR-100 (FractalDB-10k 77.3 vs. Places-365 76.9) and ImageNet (FractalDB-10k 71.5 vs. Places-365 71.4). The ImageNet-1k pre-trained model is much better than our proposed method on fine-tuning datasets such as C100 and VOC12 since these datasets contain similar categories such as animals and tools.

\begin{figure}[t]
\centering
   \includegraphics[width=0.95\linewidth]{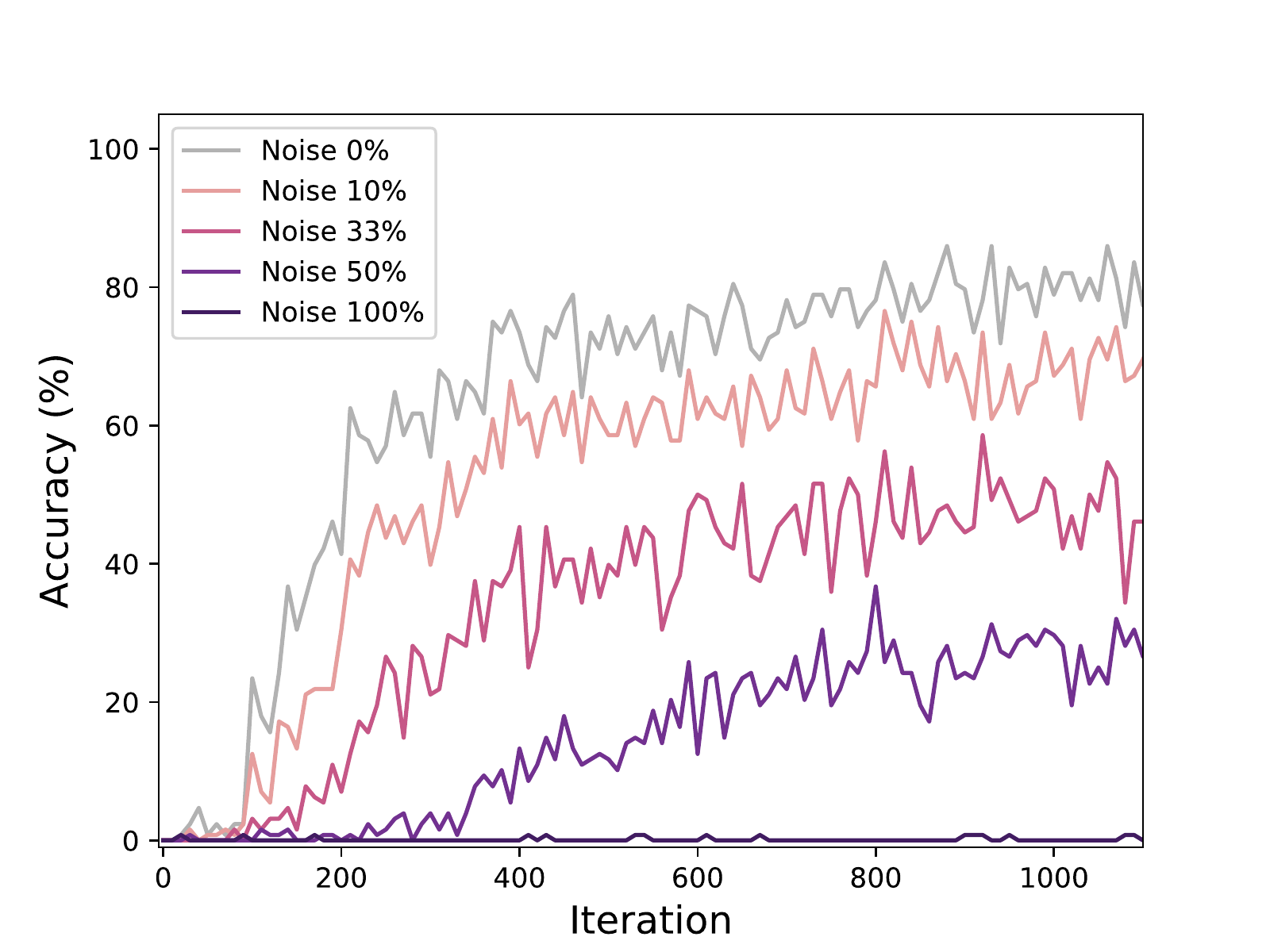}
   \caption{Noise and accuracy.}
\label{fig:acc_labelnoise}
\end{figure}

\begin{table*}[t]
\begin{center}
\caption{The classification accuracies of the FractalDB-1k/10k (F1k/F10k) and DeepCluster-10k (DC-10k). Mtd/PT Img means Method and Pre-trained images.}
\label{tab:comparison_fddb}
\begin{tabular}{lc|cccccc} \hline
    Dataset & Category$^{(classification) \%}$ \\
    \hline \hline
     Mtd & PT Img & C10 & C100 & IN1k & P365 & VOC12 & OG \\ \hline\hline
     DC-10k & Natural & 89.9 & 66.9 & 66.2 & 51.2 & 67.5 & 15.2 \\ 
     DC-10k & Formula & 83.1 & 57.0 & 65.3 & \textbf{53.4} & 60.4 & 15.3 \\ 
     \hline
     \rowcolor{gray!20}F1k & Formula & 93.4 & 75.7 & 70.3 & 49.5 & 58.9 & 20.9 \\
     \rowcolor{gray!20}F10k & Formula & \textbf{94.1} & \textbf{77.3} & \textbf{71.5} & 50.8 & \textbf{73.6} & \textbf{29.2} \\
      \hline
\end{tabular}
\end{center}
\end{table*}

\subsection{Additional experiments}
\label{sec:additionalexperiments}
We also validated the proposed framework in terms of (i) category assignment, (ii) convergence speed, (iii) freezing parameters in fine-tuning, (iv) comparison to other formula-driven image datasets, (v) recognized category analysis and (vi) visualization of first convolutional filters and attention maps.

\textbf{(i) Category assignment (see Figure~\ref{fig:acc_labelnoise} and Table~\ref{tab:comparison_fddb}).} At the beginning, we validated whether the optimization can be successfully performed using the proposed FractalDB. Figure~\ref{fig:acc_labelnoise} show the transitioned pre-training accuracies with several rates of label noise. We randomly replaced the category labels. Here, 0\% and 100\% noise indicate normal training and fully randomized training, respectively. According to the results on FractalDB-1k, a CNN model can successfully classify fractal images, which are defined by iterated functions. Moreover, well-defined categories with a balanced pixel rate allow optimization on FractalDB. 
When fully randomized labels were assigned in FractalDB training, the architecture could not correct any images and the loss value was static (the accuracies are 0\% at almost times). According to the result, we confirmed that the effect of the fractal category is reliable enough to train the image patterns.

Moreover, we used the DeepCluster-10k to automatically assign categories to the FractalDB. 
Table~\ref{tab:comparison_fddb} indicates the comparison between category assignment with DeepCluster-10k (k-means) and FractalDB-1k/10k (IFS). We confirm that the DeepCluster-10k cannot successfully assign a category to fractal images. The gaps between IFS and k-means assignments are \{11.0, 20.3, 13.2\} on \{C10, C100, VOC12\}. This obviously indicates that our formula-driven image generation through the principle of IFS and the parameters in equation (2) works well compared to the DeepCluster-10k.

\begin{table}[t]
  \begin{minipage}[h]{.48\textwidth}
    \caption{Freezing parameters.}
    \begin{center}
      \begin{tabular}{l|cccc} \hline
        Freezing layer(s) & C10 & C100 & IN100 & P30 \\
        \hline \hline
        Fine-tuning & 93.4 & 75.7 & 82.7 & 75.9 \\
        \hline
        Conv1 & 92.3 & 72.2 & 77.9 & 74.3 \\
        Conv1--2 &  92.0 & 72.0 & 77.5 & 72.9 \\
        Conv1--3 & 89.3 & 68.0 & 71.0 & 68.5 \\
        Conv1--4 & 82.7 & 56.2 & 55.0 & 58.3 \\
        Conv1--5 & 49.4 & 24.7 & 21.2 & 31.4 \\
        \hline
      \end{tabular}
    \end{center}
    \label{tab:freeze}
  \end{minipage}
  \hfill
  \begin{minipage}[h]{.48\textwidth}
    \caption{Other formula-driven image datasets with a Bezier curves and Perlin noise.}
    \begin{center}
      \begin{tabular}{l|cccc} \hline
        Pre-training & C10 & C100 & IN100 & P30 \\
        \hline \hline
        Scratch & 87.6 & 60.6 & 75.3 & 70.3 \\
        Bezier-144 & 87.6 & 62.5 & 72.7 & 73.5 \\
        Bezier-1024 & 89.7 & 68.1 & 73.0 & 73.6 \\
        Perlin-100 & 90.9 & 70.2 & 73.0 & 73.3 \\
        Perlin-1296 & 90.4 & 71.1 & 79.7 & 74.2 \\
        \hline
        \rowcolor{gray!20}FractalDB-1k & \textbf{93.4} & \textbf{75.7} &  \textbf{82.7} & \textbf{75.9} \\
        \hline
      \end{tabular}
    \end{center}
    \label{tab:bezie_perlin}
  \end{minipage}
\end{table}

\begin{table*}[t]
\begin{center}
\caption{Performance rates in which FractalDB was better than the ImageNet pre-trained model on C10/C100/IN100/P30 fine-tuning.}
\begin{tabular}{l|l} \hline
    Dataset & Category$^{(classification) \%}$ \\
    \hline \hline
    C10 & -- \\ \hline
    C100 & bee$^{(89)}$, chair$^{(92)}$, keyboard$^{(95)}$, maple tree$^{(72)}$, motor cycle$^{(99)}$, \\
         & orchid$^{(92)}$, pine tree$^{(70)}$ \\ \hline
    IN100 & Kerry blue terrier$^{(88)}$, marmot$^{(92)}$, giant panda$^{(92)}$, television$^{(80)}$, \\
         & dough$^{(64)}$, valley$^{(94)}$ \\ \hline
    P30 & cliff$^{(64)}$, mountain$^{(40)}$, skyscrape$^{(85)}$, tundra$^{(79)}$ \\ \hline
\end{tabular}
\label{tab:bettercategory}
\end{center}
\end{table*}

\begin{figure*}[t]
\centering
\subfigure[ImageNet] {\includegraphics[width=0.18\linewidth]{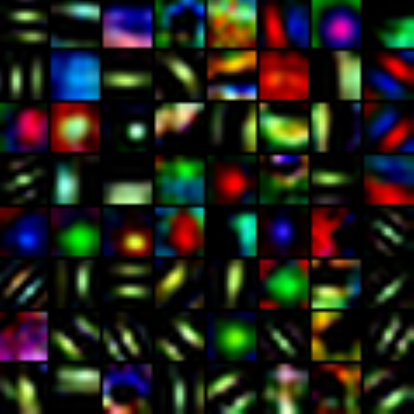}
\label{fig:ImageNet_pretrain}}
\subfigure[Places365] {\includegraphics[width=0.18\linewidth]{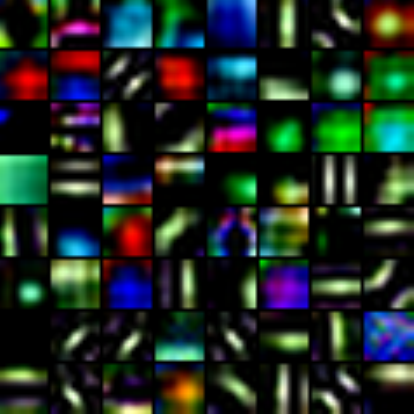}
\label{fig:Places365_pretrain}}
\subfigure[Fractal-1K]{\includegraphics[width=0.18\linewidth]{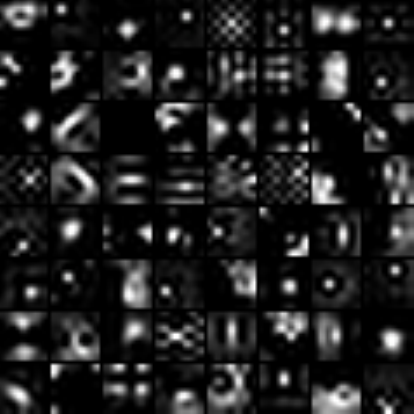}
\label{fig:fractaldb-rs1K_pretrain}}
\subfigure[Fractal-10K]{\includegraphics[width=0.18\linewidth]{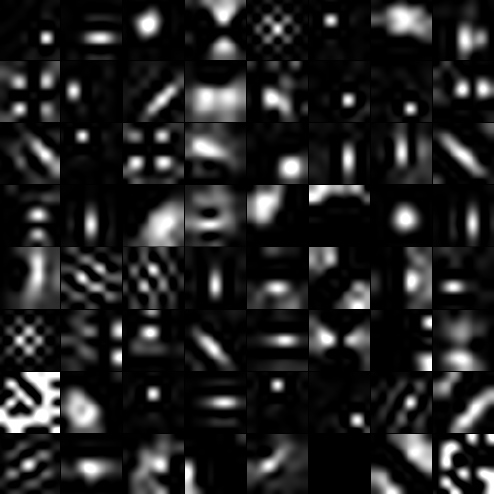}
\label{fig:fractaldb-rs10K_pretrain}}
\subfigure[DC-10k]{\includegraphics[width=0.18\linewidth]{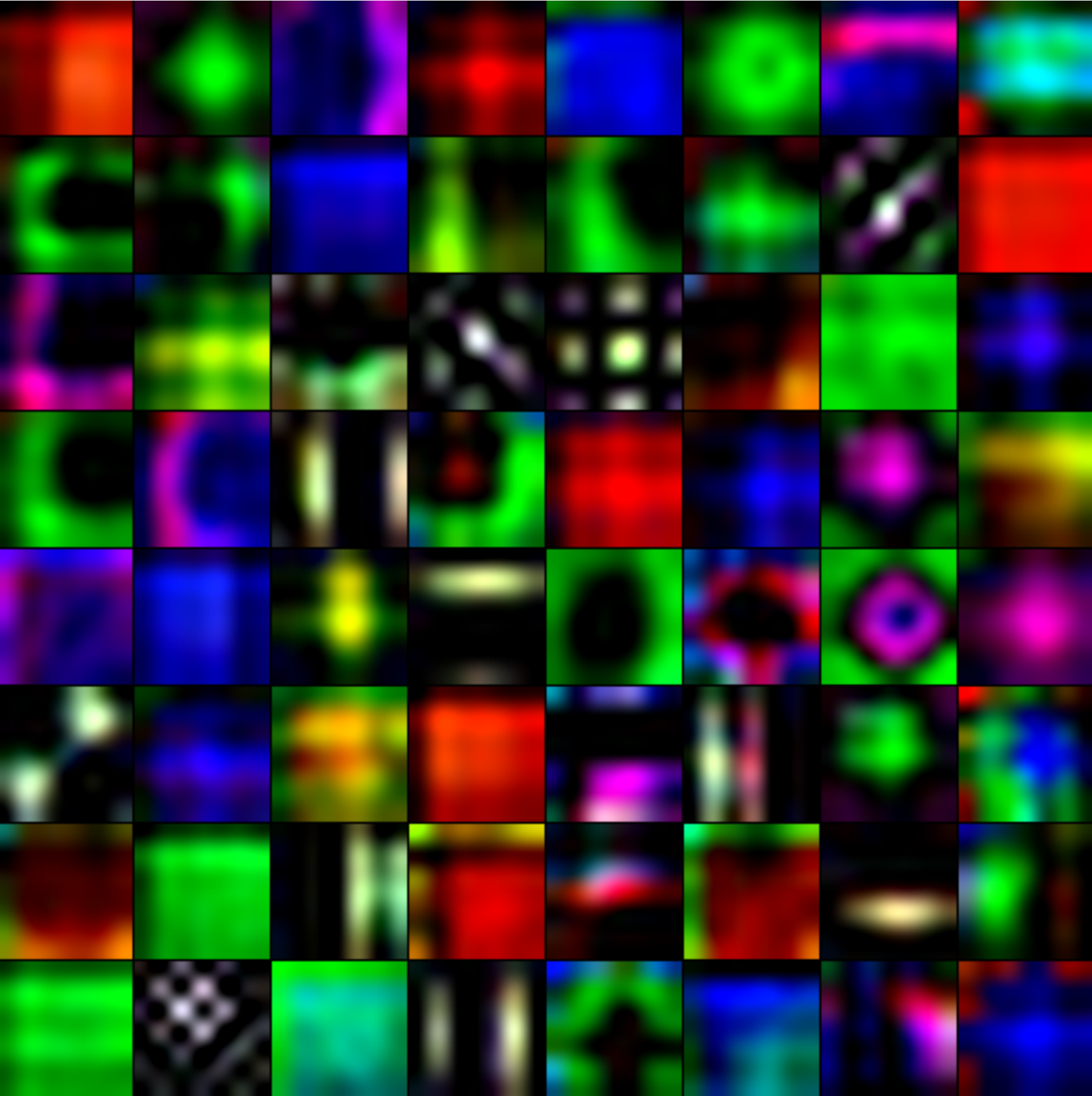}
\label{fig:deepcluster_conv1}}
\vspace{-0pt}
\\
\subfigure[Heatmaps with Grad-CAM. (Left) Input image. (Center-left) Activated heatmaps with ImageNet-1k pre-trained ResNet-50. (Center) Activated heatmaps with Places-365 pre-trained ResNet-50. (Center-Right, Right) Activated heatmaps with FractalDB-1K/10k pre-trained ResNet-50.]{\includegraphics[width=0.98\linewidth]{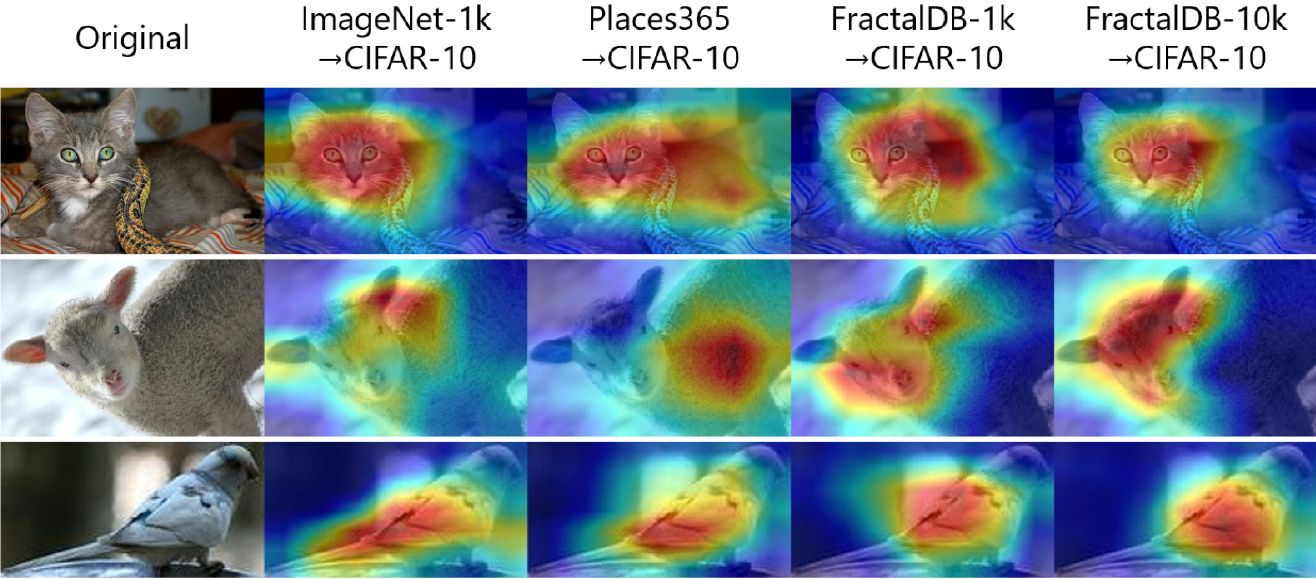}
\label{fig:grad_cam}}
\caption{Visualization results: (a)--(e) show the activation of the 1st convolutional layer on ResNet-50, and (f) illustrates attentions with Grad-CAM~\cite{Selvaraju_2017_ICCV}.
}
\label{fig:vis}
\end{figure*}

\textbf{(ii) Convergence speed (see Figure~\ref{fig:acc_compare}).} The transitioned pre-training accuracies values in FractalDB are similar to those of ImageNet pre-trained model and much faster than scratch from random parameters (Figure~\ref{fig:acc_compare}). We validated the convergence speed in fine-tuning on C10. As the result of pre-training with FractalDB-1k, we accelerated the convergence speed in fine-tuning which is similar to the ImageNet pre-trained model.

\textbf{(iii) Freezing parameters in fine-tuning (see Table~\ref{tab:freeze}).} Although full-parameter fine-tuning is better, conv1 and 2 acquired a highly accurate image representation (Table~\ref{tab:freeze}). Freezing the conv1 layer provided only a -1.4 (92.0 vs. 93.4) or -2.8 (72.9 vs. 75.7) decrease from fine-tuning on C10 and C100, respectively. Comparing to other results, such as those for conv1--4/5 freezing, the bottom layer tended to train a better representation.

\textbf{(iv) Comparison to other formula-driven image datasets (see Table~\ref{tab:bezie_perlin}).} At this moment, the proposed FractalDB-1k/10k are better than other formula-driven image datasets. We assigned Perlin noise~\cite{PerlinToG2002} and Bezier curve~\cite{beziercurve} to generate image patterns and their categories just as FractalDB made the dataset (see the supplementary material for detailed dataset creation of the Bezier curve and Perlin noise). We confirmed that Perlin noise and the Bezier curve are also beneficial in making a pre-trained model that achieved better rates than scratch training. However, the proposed FractalDB is better than these approaches (Table~\ref{tab:bezie_perlin}). For a fairer comparison, we cite a similar \#category in the formula-driven image datasets, namely FractalDB-1k (total \#image: 1M), Bezier-1024 (1.024M) and Perlin-1296 (1.296M). The significantly improved rates are +3.0 (FractalDB-1k 93.4 vs. Perlin-1296 90.4) on C10, +4.6 (FractalDB-10k 75.7 vs. Perlin-1296 71.1) on C100, +3.0 (FractalDB-1k 82.7 vs. Perlin-1296 79.7) on IN100, and +1.7 (FractalDB-1k 75.9 vs. Perlin-1296 74.2) on P30.

\textbf{(v) Recognized category analysis (see Table~\ref{tab:bettercategory}).} We investigated Which categories are better recognized by the FractalDB pre-trained model compared to the ImageNet pre-trained model. Table~\ref{tab:bettercategory} shows the category names and the classification rates. The FractalDB pre-trained model tends to be better when an image contains recursive patterns (e.g., a keyboard, maple trees).


\textbf{(vi) Visualization of first convolutional filters (see Figures~\ref{fig:vis}(a--e)) and attention maps (see Figure~\ref{fig:grad_cam}).} We visualized first convolutional filters and Grad-CAM~\cite{Selvaraju_2017_ICCV} with pre-trained ResNet-50. As seen in ImageNet-1k/Places-365/DeepCluster-10k (Figures~\ref{fig:ImageNet_pretrain}, \ref{fig:Places365_pretrain} and \ref{fig:deepcluster_conv1}) and FractalDB-1k/10k pre-training (Figures~\ref{fig:fractaldb-rs1K_pretrain} and \ref{fig:fractaldb-rs10K_pretrain}), our pre-trained models obviously generate different feature representations from conventional natural image datasets. Based on the experimental results, we confirmed that the proposed FractalDB successfully pre-trained a CNN model without any natural images even though the convolutional basis filters are different from the natural image pre-training with ImageNet-1k/DeepCluster-10k.

The pre-trained models with Grad-CAM can generate heatmaps fine-tuned on C10 dataset. According to the center-right and right in Figure~\ref{fig:grad_cam}, the FractalDB-1k/10k also look at the objects.

\section{Discussion and Conclusion}

We achieved the framework of \textit{pre-training without natural images} through formula-driven image projection based on fractals. We successfully pre-trained models on FractalDB and fine-tuned the models on several representative datasets, including CIFAR-10/100, ImageNet, Places and Pascal VOC. The performance rates were higher than those of models trained from scratch and some supervised/self-supervised learning methods. 
Here, we summarize our observations through exploration as follows.

\textbf{Towards a better pre-trained dataset.} The proposed FractalDB pre-trained model partially outperformed ImageNet-1k/Places365 pre-trained models, e.g., FractalDB-10k 77.3 vs. Places-365 76.9 on CIFAR-100, FractalDB-10k 50.8 vs. ImageNet-1k 50.3 on Places-365. If we could improve the transfer accuracy of the pre-training without natural images, then the ImageNet dataset and the pre-trained model may be replaced so as to protect fairness, preserve privacy, and decrease annotation labor. Recently, for examples, 80M Tiny Images\footnote{\url{https://groups.csail.mit.edu/vision/TinyImages/}} and ImageNet (human-related categories)\footnote{\url{http://image-net.org/update-sep-17-2019}} have been withdrawn the publicly available images.


\textbf{Are fractals a good rendering formula?} We are looking for better mathematically generated image patterns and their categories. We confirmed that FractalDB is better than datasets based on the Bezier curve and Perlin Noise in the context of pre-trained model (see Table~\ref{tab:bezie_perlin}). Moreover, the proposed FractalDB can generate a good set of categories, e.g., the fact that the training accuracy decreased depending on the label noise (see Figures~\ref{fig:acc_labelnoise}) and the formula-driven image generation is better than DeepCluster-10k in the most cases, as a method for category assignment (see Table~\ref{tab:comparison_fddb}) show how the fractal categories worked well. 

\textbf{A different image representation from human annotated datasets.} The visual patterns pre-trained by FractalDB acquire a unique feature in a different way from ImageNet-1k (see Figure~\ref{fig:vis}). In the future, steerable pre-training may be available depending on the fine-tuning task. Through our experiments, we confirm that a pre-trained dataset configuration should be adjusted. We hope that the proposed pre-training framework will suit a broader range of tasks, e.g., object detection and semantic segmentation, and will become a flexibly generated pre-training dataset.

\section*{Acknowledgement}
\begin{itemize}
    \item This work was supported by JSPS KAKENHI Grant Number JP19H01134.
    \item Computational resource of AI Bridging Cloud Infrastructure (ABCI) provided by National Institute of Advanced Industrial Science and Technology (AIST) was used. 

\end{itemize}

%
%


%
%

\bibliographystyle{spmpsci}
\bibliography{egbib}

\end{document}